# Calibrated Unsupervised Anomaly Detection in Multivariate Time-series using Reinforcement Learning

Saba Sanami and Amir G. Aghdam

*Abstract*—This paper investigates unsupervised anomaly detection in multivariate time-series data using reinforcement learning (RL) in the latent space of an autoencoder. A significant challenge is the limited availability of anomalous data, often leading to misclassifying anomalies as normal events, thus raising false negatives. RL can help overcome this limitation by promoting exploration and balancing exploitation during training, effectively preventing overfitting. Wavelet analysis is also utilized to enhance anomaly detection, enabling time-series data decomposition into both time and frequency domains. This approach captures anomalies at multiple resolutions, with wavelet coefficients extracted to detect both sudden and subtle shifts in the data, thereby refining the anomaly detection process. We calibrate the decision boundary by generating synthetic anomalies and embedding a supervised framework within the model. This supervised element aids the unsupervised learning process by fine-tuning the decision boundary and increasing the model's capacity to distinguish between normal and anomalous patterns effectively.

## I. INTRODUCTION

Anomaly detection is the process of identifying patterns in data that deviate from expected norms or behaviors. These anomalies can indicate critical events such as system failures, fraud, or unforeseen process changes. In time-series data, anomalies can appear as individual data points, abnormal sequences within a context, or unusual collective behavior over time. Different anomaly detection techniques have been developed across various domains, including industrial monitoring, finance, cybersecurity, and healthcare, where timely identification of abnormal events can prevent damage, reduce risk, and optimize operations. With the increasing complexity and volume of data, especially in some emerging applications, anomaly detection has become a key tool for maintaining system integrity and improving decision-making in automated and intelligent systems [1], [2].

There is a wide range of anomaly detection methods for different applications. Some use simple statistical models, while others rely on advanced learning techniques. Traditional statistical methods, such as autoregressive integrated moving average (ARIMA) models [3], [4], can capture temporal dependencies and identify anomalies as deviations from learned patterns. However, these methods often struggle with non-linearity and complexity in real-world data. Machine learning-based approaches, such as clustering and classification, have been adopted to address these challenges [5], [6], [7]. With the advent of Industry 4.0 and the exponential growth of data in various industrial applications, many methods have focused on deep learning, including long short-term memory (LSTM) networks and autoencoders, which have shown potential in learning complex temporal patterns and detecting anomalies [8], [9], [10], [11]. Furthermore, hybrid models that integrate several networks have also been explored to improve the accuracy of detection mechanisms [12], [13].

Anomaly detection using deep learning techniques can be performed using different types of supervision depending on the availability of labeled data. Supervised anomaly detection requires a labeled dataset containing both normal and anomalous examples. This approach enables the model to learn the characteristics of both classes during training, making it highly effective but reliant on the scarce availability of labeled anomalies. Semi-supervised anomaly detection, on the other hand, assumes access to a large amount of unlabeled data and a small set of labeled normal data. The model is designed to recognize normal behavior, with deviations indicating anomalies. This method is particularly useful when anomalies are rare or not well-defined. Unsupervised anomaly detection, the most common approach due to the scarcity of labeled anomalies, does not require labeled data. Instead, it identifies anomalies as data points that deviate significantly from the majority of the data distribution. This category encompasses clustering-based and reconstruction-based methods [14], [15], [16].

Unsupervised anomaly detection methods, while practical due to the lack of labeled abnormal data, encounter several significant challenges. One major issue is anomaly contamination, where abnormal instances are inadvertently included in the training data, making it difficult for models to distinguish between normal and abnormal patterns [17]. Another challenge is the data imbalance, as anomalies are rare, leading to models that can be biased toward classifying data as normal, increasing the risk of false negatives. Additionally, dynamic data distributions or concept drift can affect model performance, as the normal behavior of a system can change over time, making it difficult for models trained on static data to remain accurate.

While several papers have aimed at improving the accuracy of unsupervised anomaly detection, only a few have focused on calibrating the decision boundary in multivariate time-series data. This work contributes to addressing these challenges by

- Implementing a reinforcement learning (RL) agent that learns to adjust the decision boundary dynamically based on feedback from detection performance. The wavelet-based features are processed through a deep learning model to extract latent features, which the RL agent uses to make decisions about boundary calibration.
- To assist the unsupervised learning process, we generate synthetic anomalies and incorporate them into a supervised framework. This approach provides reference points

Saba Sanami and Amir G. Aghdam are with the Department of Electrical and Computer Engineering, Concordia Univerity, Montreal, QC, Canada. Emails: saba.sanami@mail.concordia.ca, amir.aghdam@concordia.ca.

for the RL agent, aiding in refining the decision boundary and enhancing the model's ability to distinguish between normal and abnormal patterns.

The rest of the paper is organized as follows. Section II presents the problem formulation. The proposed model is elaborated in Section III. The algorithm is evaluated through multivariate time series data in an aero-engine application in Section IV, and finally Section V concludes the paper.

## II. PROBLEM FORMULATION AND PRELIMINARIES

Consider a multivariate time series represented by $\mathbf{X} = [\mathbf{x}_1, \mathbf{x}_2, \ldots, \mathbf{x}_T]$, where $T$ is the total number of time steps. Each observation $\mathbf{x}_t \in R^D$ at time $t$ consists of measurements from $D$ different variables or sensors. To effectively capture both temporal dependencies and the relationships among variables, we employ a sliding window approach to segment the time series into overlapping subsequences. Specifically, we define window length $L$ as the number of consecutive time steps included in each subsequence, and the stride $\Delta$, the step size by which the window moves along the time series.

Using the parameters introduced above, we generate a set of subsequences $\mathcal{S} = \{\mathbf{S}_1, \mathbf{S}_2, \ldots, \mathbf{S}_N\}$, where each subsequence $\mathbf{S}_n$ is defined as

$$\mathbf{S}_n = [\mathbf{x}_{t_n}, \mathbf{x}_{t_n+1}, \ldots, \mathbf{x}_{t_n+L-1}].$$

To enhance the feature representation of each subsequence, we apply wavelet analysis to each subsequence $\mathbf{S_n}$, as follows

$$\mathbf{w_n} = W(\mathbf{S_n}). \tag{1}$$

We then apply the proposed anomaly detection method to $\mathbf{w_n}$ to classify normal data and anomalies by leveraging both the time and frequency information captured in these coefficients.

### A. Wavelet-Based Feature Extraction

Wavelet analysis is used for analyzing time-series data, particularly when the data exhibits non-stationary characteristics or contains features at multiple scales. Unlike traditional time-domain analysis, which examines data purely as a function of time, wavelet analysis decomposes a signal into both time and frequency components. This representation enables the capture of localized frequency changes over time, providing a more comprehensive understanding of the underlying patterns within the data. Moreover, wavelet transforms can effectively separate signal components from noise, enhancing the ability to detect meaningful patterns and anomalies. In systems with multiple operating modes or dynamic behaviors, wavelet analysis can isolate mode-specific features.

A wavelet is a finite-duration oscillatory function with a zero average value. Wavelet transforms entail convolving the original signal with scaled and shifted versions of a mother wavelet function. For each subsequence $\mathbf{S}_n$, the discrete wavelet transform (DWT) decomposes the signal into approximation and detail coefficients at multiple resolutions. The DWT applies a low-pass filter $h[n]$ and a high-pass filter $g[n]$, followed by downsampling to extract the coefficients.

The approximation coefficients (low-frequency components) are given by [18]

$$A_j[n] = \sum_k \mathbf{S}_n[k] \cdot h[2n - k],$$

and the detail coefficients (high-frequency components) are as follows

$$D_j[n] = \sum_k \mathbf{S}_n[k] \cdot g[2n - k],$$

The above two coefficients are at level $j$, and $h[n]$ and $g[n]$ are the filters applied to the subsequence $\mathbf{S}_n$. Following sequence generation, each variable in it undergoes DWT, and the resulting coefficients (which represent different wavelet components) are flattened and padded to a consistent size. Finally, these coefficients are stacked side by side for each variable, creating an image $\mathbf{w_n}$ where rows represent different frequency components and columns represent different features.

## III. MODEL OVERVIEW

The proposed model for unsupervised anomaly detection in multivariate time-series data integrates a convolutional neural network autoencoder (CNN-AE) with RL to create a structured latent space that effectively distinguishes between normal and anomalous data. Using an Actor-Critic approach, the CNN-AE, acting as the actor, is trained to minimize reconstruction loss and maximize rewards from the critic, which evaluates how well the latent space separates different data classes. Since we add synthetic anomaly data, the decision boundary can be calibrated according to these added anomalies. This combination of reconstruction and reward signals encourages distinct clustering of anomaly types, thereby improving the interpretability and accuracy of anomaly detection. We explain the details of each component in the sequel.

### A. Initial Anomaly Detection via Reconstruction Error

The wavelet-transformed coefficients $\mathbf{w}_n$ serve as inputs to a CNN-AE, designed to learn compressed representations of the data. The encoder maps input $\mathbf{w}_n$ to a latent representation $\mathbf{z}_n$. The decoder reconstructs the input from the latent representation.

The autoencoder is trained to minimize the reconstruction loss

$$\mathcal{L}_{\text{recon}} = \frac{1}{N} \sum_{n=1}^{N} ||\mathbf{w}_n - \hat{\mathbf{w}}_n||_2^2, \tag{2}$$

where $\mathbf{w}_n$ represents the input wavelet-transformed coefficients, and $\hat{\mathbf{w}}_n$ is the reconstructed output from the decoder. Training is performed using only data assumed to be normal, enabling the model to learn the underlying patterns of normal behavior [19].

After training, the CNN-AE is used to reconstruct the inputs, and the reconstruction error for each subsequence is computed as

$$e_n = ||\mathbf{w}_n - \hat{\mathbf{w}}_n||_2^2. \tag{3}$$

An initial decision boundary (threshold) $\theta_0$ is established based on the distribution of reconstruction errors from the normal data. Observations with reconstruction errors exceeding this threshold are considered potential anomalies.

### B. Reinforcement Learning for Dynamic Decision Boundary Calibration

To enhance the separation of different anomaly types in the latent space and the adaptability of the model, we introduce an RL agent that dynamically adjusts the encoder parameters and the decision boundary. The RL agent interacts with the autoencoder to promote clustering of anomalies and separation between two clusters in the latent space, thereby improving the distinction between normal and anomalous patterns. We generate synthetic anomalies to guide the RL agent in learning an effective decision boundary.

We formulate the problem as a Markov decision process (MDP), characterized by a tuple $(\mathcal{S}, \mathcal{A}, P, R, \gamma)$, where $\mathcal{S}$ is the set of states, $\mathcal{A}$ is the set of actions, $P(s_{n+1}|s_n, a_n)$ is the state transition probability, $R(s_n, a_n)$ is the reward function, and $\gamma \in [0, 1]$ is the discount factor. The state at time step $n$ is defined as the latent representation of the input subsequence

$$s_n = \mathbf{z}_n = f_{\text{enc}}(\mathbf{w}_n; \theta_{\text{enc}}), \qquad (4)$$

where $f_{\text{enc}}$ is the encoder function parameterized by $\theta_{\text{enc}}$ and $\mathbf{w}_n$. The action $a_n$ consists of adjustments to $\theta_{\text{enc}}$ and the decision boundary $\theta_n$. Specifically, the action can be represented as

$$a_n = (\Delta\theta_{\text{enc}}, \Delta\theta_n), \qquad (5)$$

where $\Delta\theta_{\text{enc}}$ denotes an update to the encoder parameters, and $\Delta\theta_n$ is an adjustment to the decision boundary. The policy defines the probability of selecting action $a_n$ given state $s_n$. The next state $s_{n+1}$ depends on the current state and the action taken

$$s_{n+1} = f_{\text{enc}}(\mathbf{w}_{n+1}; \theta_{\text{enc}} + \Delta\theta_{\text{enc}}). \qquad (6)$$

The reward function $R(s_n, a_n)$ is designed to achieve maximum separation between clusters in the latent space corresponding to anomalies and normal data. It consists of two components, expressed below

$$R(s_n, a_n) = R_{\text{sep}}(s_n, a_n) + R_{\text{acc}}(s_n, a_n), \qquad (7)$$

where $R_{\text{sep}}(s_n, a_n)$ is the separation reward, promoting maximum distance between clusters of different classes in latent space, and $R_{\text{acc}}(s_n, a_n)$ is the accuracy reward, encouraging correct classification of data points based on the decision boundary. We define the separation between the centroids of the normal data $\boldsymbol{\mu}_0$ and the synthetic anomalies $\boldsymbol{\mu}_1$ in the latent space

$$R_{\text{sep}}(s_n, a_n) = \|\boldsymbol{\mu}_0 - \boldsymbol{\mu}_1\|_2^2$$

.

By applying Monte Carlo sampling to generate multiple reconstructions, we estimate the uncertainty for each sample's reconstruction. We consider samples with low uncertainty in reconstruction as normal, whereas synthetic anomalies are explicitly labeled as abnormal. This allows us to define $R_{acc}$ based on the classification accuracy of the synthetic anomalies and normal data. Since we have labels for the synthetic anomalies ($y_n = 1$) and by this assumption that the rest are normal ($y_n = 0$), we define $R_{\text{acc}}$ as

$$R_{\text{acc}}(s_n, a_n) = \begin{cases} +1, & \text{if} \quad \hat{y}_n = y_n \\ -1, & \text{if} \quad \hat{y}_n \neq y_n \end{cases} \qquad (8)$$

where $\hat{y}_n$ is the predicted label based on the adjusted decision boundary $\theta_n$

$$\hat{y}_n = \begin{cases} 0, & \text{if } e_n \leq \theta_n \\ 1, & \text{if } e_n > \theta_n \end{cases}$$

The RL agent aims to learn an optimal Q-function $Q^*(s, a)$ that maximizes the expected cumulative reward, defined as

$$Q(s_n, a_n) = E\left[\sum_{n=0}^{\infty} \gamma^n R(s_n, a_n) \bigg| s_0 = s, a_0 = a\right]. \qquad (9)$$

The Q-function $Q(s_n, a_n)$ represents the expected cumulative reward starting from state $s_n$, taking action $a_n$, and following the optimal policy thereafter.

We employ Q-learning to train the RL agent. The Q-learning update rule is

$$\begin{aligned} Q(s_n, a_n) = Q(s_n, a_n) + \alpha\Big[&R(s_n, a_n)+ \\ &\gamma \max_{a'} Q(s_{n+1}, a') - Q(s_n, a_n)\Big], \end{aligned} \qquad (10)$$

where $\alpha$ is the learning rate, and $a'$ denotes the possible actions that the agent could take from the next state $s_{n+1}$. The term $\max_{a'} Q(s_{n+1}, a')$ denotes the maximum Q-value over all possible actions in state $s_{n+1}$, reflecting the expected future reward for the best action from that state.

The overall objective function of our model now combines the reconstruction loss from the autoencoder and the RL agent's goal of maximizing the cumulative reward

$$\mathcal{L}_{\text{total}} = \mathcal{L}_{\text{recon}} - \eta Q(s_n, a_n) \qquad (11)$$

where $\eta$ is a weighting factor that balances the reconstruction loss and the RL agent's objective. The negative sign indicates that maximizing $Q(s_n, a_n)$ contributes to minimizing the total loss.

By incorporating the RL agent into the training process, we encourage the autoencoder to produce latent representations that are well-clustered according to their classes. The RL agent dynamically adjusts the encoder parameters and the decision boundary to maximize the separation between clusters, enhancing the model's ability to distinguish between normal data and various types of anomalies. The process is summarized in Algorithm 1.

## Algorithm 1 Training the RL Agent for Dynamic Decision Boundary Calibration

1: **Initialization:**
2: Set initial decision boundary $\theta_0$ and initialize Q-function $Q(s, a)$.
3: Generate synthetic anomalies by perturbing normal data and assign labels $y_n$.
4: Initialize encoder parameters $\theta_{\text{enc}}$ and decoder parameters $\theta_{\text{dec}}$.
5: **for** each episode **do**
6:    **for** each subsequence $\mathbf{w}_n$ **do**
7:       **Encode** input to obtain latent representation:
$$\mathbf{z}_n = f_{\text{enc}}(\mathbf{w}_n; \theta_{\text{enc}})$$
8:       **Observe** state $s_n = \mathbf{z}_n$.
9:       **Select** action $a_n$ based on $Q(s_n, a_n)$.
10:      **Apply** action $\theta_n \leftarrow \theta_n + \Delta\theta_n$
11:      **Compute** reconstruction error $e_n$
12:      **Predict** label $\hat{y}_n$
13:      **Compute** rewards $R_{\text{sep}}$ and $R_{\text{acc}}$
14:      **Receive** total reward $R(s_n, a_n) = R_{\text{sep}} + R_{\text{acc}}$
15:      **Observe** next state $s_{n+1} = s_n$.
16:      **Update** Q-function:
17:      **Store** transition $(s_n, a_n, R(s_n, a_n), s_{n+1})$.
18:    **end for**
19:    **Update** encoder and decoder parameters $\theta_{\text{enc}}, \theta_{\text{dec}}$ by minimizing the total loss $\mathcal{L}_{\text{total}} = \mathcal{L}_{\text{recon}} - \eta \sum_n Q(s_n, a_n)$
20: **end for**

## IV. EXPERIMENT

**Example 1**- The original dataset used in this example, known as C-MAPSS [20], captures multiple critical aero-engine parameters, including temperatures, pressures, rotational speeds, and flow rates across key components such as the low-pressure compressor (LPC) and high-pressure compressor (HPC). Degradation is observed towards the end of the sequence. Figure 1 presents three sensor data. For wavelet analysis, we first normalize the data and create overlapping sequences of length 10 using a sliding window with a stride of 2, allowing us to capture temporal dependencies and maintain context across consecutive sequences. Each sequence is then processed independently using the DWT with the Daubechies 1 (db1) wavelet for each feature.

The resulting DWT coefficients for all features in each sequence are combined side-by-side to create a two-dimensional image, shown in Fig. 2. Each column in this figure represents a different feature's DWT coefficients, and each row represents the frequency component index. This approach allows us to visualize the aero-engine data in a compact, image-like representation, highlighting the variations in different features across multiple frequencies. The DWT patterns show clear differences between the first and last sequences, suggesting abnormalities in the latter stages. However, without labeled data, the exact point of abnormality is uncertain, so we create synthetic abnormalities in the signals and label them as abnormal to guide the anomaly detection model.

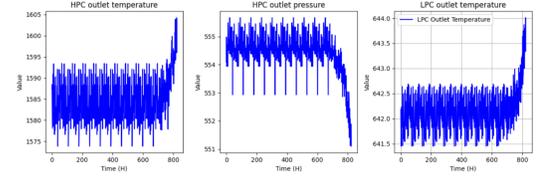

Fig. 1. Original data samples

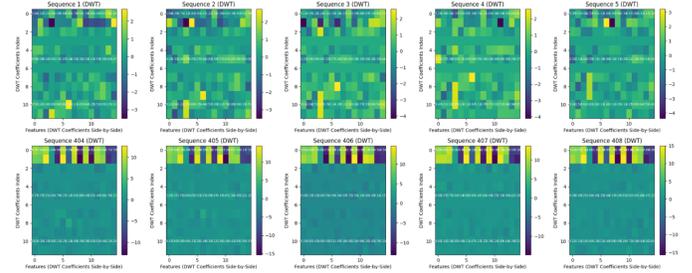

Fig. 2. DWT coefficients for first and last 5 sequences

We generate three types of anomalies: cyclic anomalies (applied to all correlated data), sudden sensor drift, and gradual sensor drift, as illustrated in Fig. 3. The cyclic anomaly simulates periodic fluctuations caused by mechanical oscillations in the aero-engine. The sudden sensor drift represents an abrupt change in sensor readings, which may indicate transient faults or sensor malfunctions. The gradual sensor drift captures a slow, progressive deviation in sensor output, often due to component wear or sensor degradation. The corresponding DWT coefficient plot, shown in Fig. 4, clearly differs from the normal signal, highlighting the changes introduced by these anomalies and their impact on the signal characteristics.

The next stage is training a CNN-AE to learn a compact representation of normal data behavior. The encoder consists of two convolutional layers (with 16 and 8 filters of size 3x3) followed by a flattening layer and a latent space representation of three dimensions. To incorporate uncertainty estimation, we add a dropout layer with a rate of 0.5 in the latent space. The decoder reconstructs the data by using a fully connected layer followed by reshaping and convolutional transpose layers. During training, we use Monte Carlo dropout to generate multiple latent space representations for each data point, allowing us to estimate uncertainty by calculating the standard deviation of these representations. We categorize data points into low or high uncertainty groups based on a threshold set at the 75th percentile of uncertainty scores.

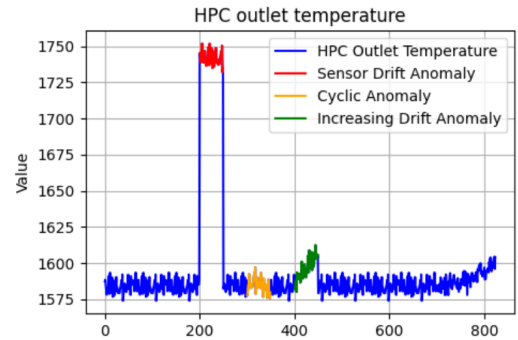

Fig. 3. HPC outlet temperature with synthetic anomalies

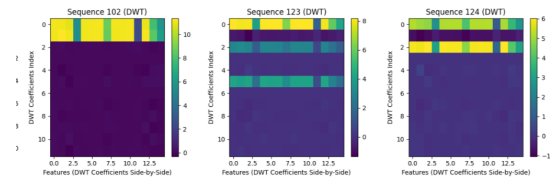

Fig. 4. DWT coefficient for synthetic anomalies sequences

Next, we implement RL to refine the decision boundary in the latent space. A Q-learning algorithm is employed, with two possible actions for the RL agent, classify data as normal or classify it as abnormal. The parameter values are given in Table 1. We initialize a Q-table with zeros, and the RL agent learns to update its policy through repeated interactions with the latent space. The agent receives a positive reward for correct classifications and a negative reward for misclassifications. High uncertainty data points are classified as abnormal, along with synthetic data, to ensure that potential anomalies are appropriately captured.

Fig. 5 represents the latent dimensions (1, 2, and 3) along three axes, and the data points are projected onto this space to identify clusters of normal and abnormal behavior. The blue dots represent the normal data points that have low uncertainty. These points are clustered closely together, indicating consistency in the data that is effectively learned by the model. The yellow circles represent real abnormal data points that were classified as high uncertainty by the model. The RL model uses these high uncertainty data points, along with the synthetic abnormalities, to refine the decision boundary and accurately separate normal and abnormal behaviors. The red X markers denote synthetic abnormal data. These synthetic anomalies are placed in the high uncertainty category to help guide the model in learning the features that distinguish normal from abnormal patterns. The RL-based decision boundary, highlighted in light red, demonstrates the model's understanding of the space that contains abnormal behavior. This decision boundary encompasses both synthetic and real abnormal data points, effectively isolating them from the cluster of normal data.

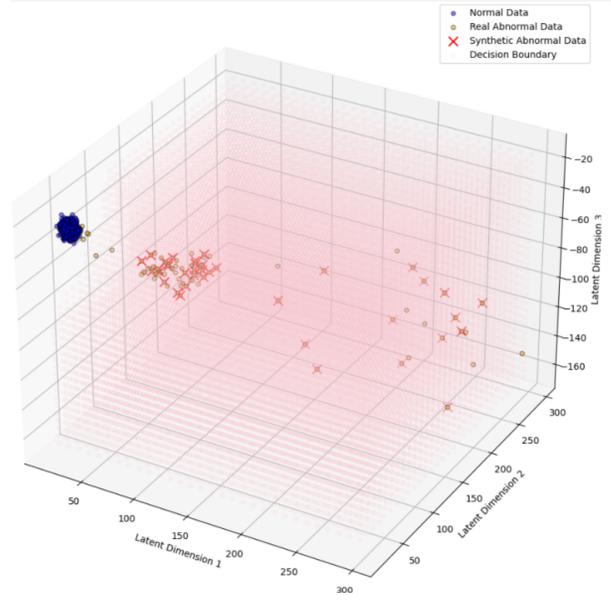

Fig. 5. Latent space representation with RL decision boundary

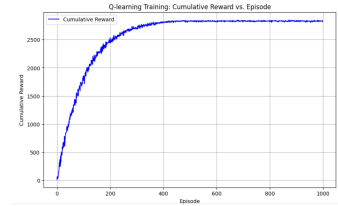

Fig. 6. Cumulative Trend for Q-Learning

TABLE I
ASSIGNED VALUES FOR RL MODEL PARAMETERS

| Parameter | Assigned Value |
| --- | --- |
| Learning rate ($\alpha$) | 0.01 |
| Discount factor ($\gamma$) | 0.95 |
| Exploration rate ($\epsilon$) | 0.1 (decayed over time) |
| Exploration decay rate | 0.99 |
| Initial decision boundary ($\theta_0$) | 0.5 |
| Reward weight for separation ($\lambda$) | 1.0 |
| Number of episode | 1000 |

Fig. 6 represents the cumulative reward for Q-learning training over 1000 episodes. Initially, the agent experiences some negative rewards as it explores its environment and tries different actions without much knowledge. However, as training progresses, we observe a significant and consistent upward trend in cumulative rewards, indicating that the agent is learning an effective policy for distinguishing between normal and abnormal data. The positive trend demonstrates that the agent is increasingly successful in taking actions that yield rewards, suggesting that it has learned to make more accurate classifications over time.

We employ several evaluation metrics to assess our anomaly detection model. Although true labels for the anomalies are not available, we devise a labeling strategy to approximate the distinction between normal and abnormal data. Specifically, instances with synthetic anomalies, high reconstruction errors, and high uncertainty in the latent space are treated as anomalies, as these often indicate deviation from learned normal patterns. Conversely, data with low reconstruction error and low uncertainty are considered normal. By using this labeling approach, we compute metrics such as precision, recall, accuracy, and F1-score, which provide a robust measure of the model's ability to discriminate between normal and abnormal behaviors effectively. We then compare the corresponding metrics with those of LSTM-AE, as shown in Table II. The proposed method consistently outperformed the LSTM-AE model in all metrics, confirming its superior performance in anomaly detection.

TABLE II
EVALUATION METRICS FOR PROPOSED METHOD AND LSTM-AE

| Metric | Proposed Method | LSTM-AE |
|---|---|---|
| Precision | 0.921 | 0.791 |
| Recall | 0.942 | 0.814 |
| Accuracy | 0.911 | 0.826 |
| F1 Score | 0.931 | 0.802 |

## V. CONCLUSIONS

In this paper, we propose an unsupervised anomaly detection framework for multivariate time-series data that leverages the RL mechanism to dynamically refine the decision boundary for autoencoder model. We employ wavelet analysis to capture both time and frequency features of the data, ensuring that subtle and transient behaviors are effectively represented. A Q-learning-based RL agent optimizes the decision boundary, aiming to improve the separation between normal and anomalous data by learning from latent space representations. To address the challenge of limited labeled data, we use synthetic anomalies, which provide supervised-like guidance to the RL agent in refining its policy. Our experimental results show that this integrated approach is effective in identifying anomalies. For future work, we aim to investigate the use of more advanced reinforcement learning algorithms, such as deep Q-networks (DQN) or policy gradient methods, to handle higher-dimensional and more complex latent spaces.